\pdfoutput=1

\documentclass[11pt]{article}

\usepackage[]{acl}
\usepackage{afterpage} 
\usepackage{rotating}
\usepackage{arabtex}
\usepackage{utf8}
\setcode{utf8}
\usepackage{tcolorbox}
\usepackage{enumitem}

\usepackage{pgfplots}
\usetikzlibrary{patterns} 
\pgfplotsset{compat=1.18} 

\usepackage{makecell}
\usepackage{graphicx}
\usepackage{times}
\usepackage{latexsym}
\usepackage{longtable}
\usepackage{multirow}
\usepackage{booktabs}
\usepackage[group-separator={,},group-minimum-digits=4]{siunitx}
\usepackage[T1]{fontenc}

\usepackage[utf8]{inputenc}

\usepackage{microtype}

\usepackage{inconsolata}
\usepackage{soul}
\usepackage{xcolor} 
\usepackage{todonotes}

\usepackage{listings}
\usepackage{xcolor}
\usepackage{amsmath}
\usepackage{subcaption}
\usepackage{fancybox}
\usepackage{tikz}

\definecolor{codegray}{gray}{0.9}

\lstdefinestyle{pythonstyle}{
    backgroundcolor=\color{codegray},   
    language=Python,
    basicstyle=\ttfamily\footnotesize,
    keywordstyle=\color{blue},
    stringstyle=\color{red},    
    breaklines=true,
    frame=single,
    keepspaces=true,
    showstringspaces=false,
}

\definecolor{lightskyblue}{rgb}{0.53, 0.81, 0.98}
\usepackage{graphicx}
\definecolor{blue}{rgb}{0,0, 0.6}
\definecolor{dkgreen}{rgb}{0,0.6,0}
\definecolor{lightgreen}{rgb}{0,0.9,0}
\definecolor{lightyellow}{rgb}{1,1,0.9}
\definecolor{gray}{rgb}{0.5,0.5,0.5}
\definecolor{mauve}{rgb}{0.58,0,0.82}
\definecolor{mauve}{rgb}{0,0,0}
\definecolor{black}{rgb}{0,0,0}
\definecolor{tri}{rgb}{.25,.88,.82}
\definecolor{lilac}{rgb}{0.85,0.64,0.85}
\definecolor{lightblue}{rgb}{0.53, 0.81, 0.98}
\definecolor{lightred}{rgb}{1.0, 0.6, 0.6}

\newcommand{\balsam}{\emph{BALSAM} }

\setcellgapes{3pt}

\title{BALSAM: A Platform for Benchmarking Arabic Large Language Models}
\author{
Rawan Al-Matham, Kareem Darwish, Raghad Al-Rasheed, Waad Alshammari,\\
\textbf{Muneera Alhoshan, Amal Almazrua, Asma Al Wazrah, Mais Alheraki, Firoj Alam,}\\
\textbf{Preslav Nakov, Norah Alzahrani, Eman alBilali, Nizar Habash, Abdelrahman El-Sheikh,}\\
\textbf{Muhammad Elmallah, Haonan Li, Hamdy Mubarak, Mohamed Anwar, Zaid Alyafeai,} \\ 
\textbf{Ahmed Abdelali, Nora Altwairesh, Maram Hasanain, Abdulmohsen Al Thubaity,}\\
\textbf{Shady Shehata, Bashar Alhafni, Injy Hamed,Go Inoue, Khalid Elmadani, Ossama Obeid, }\\
\textbf{Fatima Haouari, Tamer Elsayed, Emad Alghamdi,Khalid Almubarak, Saied Alshahrani,} \\
\textbf{Ola Aljarrah, Safa Alajlan,Areej Alshaqarawi, Maryam Alshihri, Sultana Alghurabi,} \\
\textbf{Atikah Alzeghayer, Afrah Altamimi, Abdullah Alfaifi, Abdulrahman AlOsaimy}}

\setcode{utf8}
\begin{document}
\maketitle

\begin{abstract}
The impressive advancement of Large Language Models (LLMs) in English has not been matched across all languages. In particular, LLM performance in Arabic lags behind, due to data scarcity, linguistic diversity of Arabic and its dialects, morphological complexity, etc. Progress is further hindered by the quality of Arabic benchmarks, which typically rely on static, publicly available data, lack comprehensive task coverage, or do not provide dedicated platforms with blind test sets. This makes it challenging to measure actual progress and to mitigate data contamination. Here, we aim to bridge these gaps. In particular, we introduce BALSAM, a comprehensive, community-driven benchmark aimed at advancing Arabic LLM development and evaluation. It includes 78 NLP tasks from 14 broad categories, with 52K examples divided into  37K test and 15K development, and a centralized, transparent platform for blind evaluation. We envision BALSAM as a unifying platform that sets standards and promotes collaborative research to advance Arabic LLM capabilities.
\end{list} 
\end{abstract}

\section{Introduction}

Arabic is a prominent world language with more than 400 million speakers; moreover, it is religiously significant for two billion Muslims. This has translated into significant demand for robust Arabic Natural Language Processing (NLP) systems, resulting in the development of multiple Arabic-centric Large Language Models (LLMs), such as Jais~\cite{sengupta2023jais} and Fanar~\cite{fanar2024}, and in improved Arabic support in multilingual models such as Llama~\cite{dubey2024llama}, Gemini~\cite{team2023gemini}, GPT-4o~\cite{openai2023gpt}, etc. 
Yet, despite recent progress, LLMs still underperform in Arabic compared to English. This gap stems from limited training data, the linguistic diversity of Modern Standard Arabic (MSA) and regional dialects, and Arabic's complex morphology. 

Robust benchmarking is crucial to quantify the gaps and guide future improvements in  Arabic capabilities of LLMs.
Yet, existing Arabic benchmarking initiatives, such as LAraBench~\cite{abdelali-etal-2024-larabench}, have primarily focused on standard natural language generation and understanding tasks. A more recent effort, AraGen~\cite{AraGen}, introduced a leaderboard-based framework that evaluates LLM performance across multiple dimensions, including correctness, completeness, conciseness, helpfulness, honesty, and harmlessness, in an LLM-as-a-judge setup. In parallel, several datasets have been developed to assess LLM capabilities across different dimensions: ArabicMMLU~\cite{koto-etal-2024-arabicmmlu} targets world knowledge, AraDICE~\cite{mousi-etal-2025-aradice} focuses on dialects with cognitive and cultural understanding, Palm~\cite{alwajih2025palm} addresses cultural comprehension, and \citet{ashraf-etal-2025-arabic} focus on safety.
However, existing efforts address limited LLM capabilities, lack comprehensive coverage, and have no dedicated \emph{platforms} for community collaboration. Critically, measuring progress in a consistent and reliable manner requires a standardized, community-driven framework with blind test datasets, an aspect that remains largely underdeveloped. 

Here, we aim to bridge this gap. In particular, we present the \emph{Benchmark for Arabic Language Models} (\textit{BALSAM}),\footnote{The platform is available at \url{https://benchmarks.ksaa.gov.sa}}
which is a comprehensive community-driven initiative designed to advance benchmarking efforts for Arabic LLMs.
\balsam{} includes a collection of 78 tasks across 14 categories, with a total of 52K examples divided into 37K test and 15K dev. These tasks span a wide range of natural language understanding and generation tasks, including summarization, question answering, information extraction, machine translation, and text classification, among others. 

\balsam further provides an integrated \textit{evaluation platform} featuring an Arabic LLM Leaderboard. This enables the research community to systematically assess the performance of Arabic LLMs, to monitor progress over time, and to access up-to-date benchmark results for the top-performing LLMs.
The \balsam platform goes beyond a traditional leaderboard, serving as a collaborative effort for leading academic and governmental institutions across the Middle East and beyond. Its core mission is to drive the creation of domain-specific test datasets and to establish robust benchmarks for evaluating Arabic LLMs. By promoting transparency and cooperation, \balsam aims to unify the Arabic NLP community around shared datasets and standards. Further, we investigate a variety of automated metrics and measure their correlation with human evaluation.  We show that using LLM-as-a-Judge highly correlates with human judgments while other measures such as BLEU, ROUGE, and BertScore don't.
The contributions of \balsam and this paper are summarized as follows:

\begin{itemize}[noitemsep,topsep=0pt,labelsep=.5em,leftmargin=*]
    \item \balsam is a community driven consortium that provides a centralized evaluation platform with an associated leaderboard.
    \item \balsam provide diverse dev/test sets based on 78 tasks, where the test sets are blind. 
    \item We compare the efficacy of using automated evaluations based on BLEU, ROUGE, BERTScore, and LLM-as-a-judge compared to human judgments.
\end{itemize}

\section{Related Work}

\subsection{Arabic-Centric Benchmarks}

Recent efforts have focused on benchmarking LLMs for Arabic, targeting tasks such as natural language understanding, generation, and speech processing \cite{abdelali-etal-2024-larabench, elmadany-etal-2023-orca, nagoudi-etal-2023-dolphin}. While LLMs have demonstrated remarkable capabilities across various domains, including solving graduate-level mathematical problems and passing medical examinations, 
these achievements have been predominantly assessed using English-language benchmarks. Thus, in order to evaluate and advance the performance of LLMs for Arabic, there is a critical need for the development of dedicated Arabic benchmarks. \citet{koto-etal-2024-arabicmmlu} developed ArabicMMLU, an Arabic version of the MMLU benchmark constructed from authentic school exam questions sourced from Arabic-speaking countries, without relying on translation. Similarly, \citet{mousi-etal-2025-aradice} created resources for MSA and dialectal Arabic, aiming to assess linguistic, cognitive, and cultural competencies.  \citet{alwajih2025palm} introduced datasets to evaluate the cultural and dialectal capabilities of LLMs.  
\citet{almazrouei-etal-2023-alghafa} adopted and restructured existing datasets to create benchmarks for evaluating LLMs in MSA and dialectal Arabic. 
Moreover, resources have been developed to assess domain-specific knowledge, e.g., ArabLegalEval \cite{hijazi-etal-2024-arablegaleval} focuses on legal knowledge, while Qiyas \cite{alkhalifa2024qiyasbenchmarkmeasuringchatgpt} targets mathematical reasoning. Finally, \citet{ashraf-etal-2025-arabic} developed an Arabic dataset for safety.

\subsection{English/Multilingual Benchmarks}
Several prominent benchmarks remain focused on English-centric evaluations, including MMLU \cite{hendrycksmeasuring}, HELM \cite{liang2023holisticevaluationlanguagemodels}, and BIG-bench \cite{srivastavabeyond}. MMLU is designed to assess reasoning and knowledge in real-world contexts, while HELM evaluates LLMs across a variety of metrics and scenarios. BIG-bench offers an extensive evaluation framework comprising 214 tasks, some of which include coverage of low-resource languages. Additionally, a range of multilingual benchmarks have been developed to assess model performance across diverse languages, including morphologically complex and low-resource languages such as Arabic.

\subsection{Tools and Leaderboards}

As LLMs continue to advance rapidly, it has become essential to compare their performance across various capabilities and domains. Over time, numerous tools and leaderboards have been developed to facilitate such evaluations. This includes \href{https://llmebench.qcri.org/}{LLMeBench}, a comprehensive benchmarking platform with a primary focus on Arabic NLP, speech, and multimodal tasks \cite{dalvi2023llmebench}. Moreover, tools such as \href{https://github.com/EleutherAI/lm-evaluation-harness}{LM-Evaluation-Harness}, \href{https://github.com/open-compass/opencompass}{OpenCompass}, and \href{https://github.com/bigcode-project/bigcode-evaluation-harnes}{BigCode-Evaluation-Harness} provide standardized frameworks for assessing model performance across a wide range of tasks and datasets, facilitating more robust and comprehensive comparisons, as well as signaling to LLM developers areas in which their models need improvement.  Several open-source leaderboard initiatives have emerged to benchmark Arabic language models, including the
\href{https://huggingface.co/spaces/OALL/Open-Arabic-LLM-Leaderboard}{Open Arabic LLM Leaderboard}, 
\href{https://huggingface.co/spaces/Omartificial-Intelligence-Space/Arabic-MMMLU-Leaderborad} the {Arabic-MMMLU-Leaderboard}~\cite{nacar-etal-2025-towards},
and \href{https://huggingface.co/blog/leaderboard-3c3h-aragen}{AraGen}~\cite{AraGen}.
Each of them serves a specific purpose. For example, the Arabic-MMMLU-Leaderboard is based on the MMMLU OpenAI benchmark, while AraGen focuses on a diverse set of tasks such as question answering, summarization, and reasoning.

\begin{figure}[t]
\centering
\includegraphics[width=0.5\textwidth]{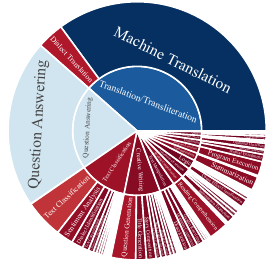}
\vspace{-0.3cm}
\caption{\label{fig:data_dist} \balsam data distribution across general categories and tasks in these categories.}
\end{figure}

\subsection{Challenges and Gaps}

Existing evaluation benchmarks rely on static, publicly available datasets, enabling rapid community assessment. Yet, as LLMs advance rapidly, static benchmarks struggle to capture their evolving capabilities. The growing size of LLMs and their increasingly extensive training data heighten the risk of test data contamination, which is difficult to detect due to opaque training data and widespread use of synthetic data \cite{dong-etal-2024-generalization}. Hence, leaderboards with rigorous contamination checks and adaptive benchmarks that reflect the latest model capabilities are needed \cite{deng2023benchmark}.

The LMSYS Chatbot Arena \cite{zheng2023judging,chiang2024chatbot} enables robust evaluation of LLMs through conversational interactions and Elo-based rankings \cite{bai2022training}, but relies on human evaluation, which is time-consuming and limits scalability \cite{luo2024arena}. The LLM-as-a-judge approach was introduced to reduce human involvement on platforms such as Chatbot Arena and MT-bench \cite{10.5555/3666122.3668142}, but it requires careful handling to avoid biases such as verbosity, position, and self-enhancement. Moreover, this method struggles with assessing reasoning and math tasks. Indeed, several popular leaderboards, including MT-bench and OpenLLM, face issues of saturation and inconsistent alignment with real-world chatbot performance \cite{luo2024arena}.

Despite significant progress in developing English benchmarks and LLM leaderboards, there remains much work to be done for languages such as Arabic. This includes the creation of new datasets to address emerging capabilities and the establishment of sustainable leaderboards that integrate human and LLM-based evaluation approaches.

\section{\balsam Dataset}


\subsection{Dataset Creation}
The \balsam benchmark is composed of 78 tasks from 14 coarse-grained categories, with a total of 52K examples divided  37K test and 15K development , and a centralized, transparent platform for blind evaluation. We made the design decision to have many datasets, but only have 10--100 test examples per dataset. For most datasets, we also have up to 50 development examples. 

Figure~\ref{fig:data_dist} shows the data distribution across general categories and tasks in these categories. We can see that the main categories are multiple-choice questions (MCQ), text generation, translation, and transliteration. 
Table~\ref{tab:phase1_dataset_statistics} and Table~\ref{tab:phase2_dataset_statistics} in the Appendix gives the complete list of tasks in \balsam along with the sizes of their development and test sets. The number of examples varies widely between tasks, with some tasks containing thousands of samples and others only a few. Figure~\ref{table:samples} in the Appendix shows sample entries from different categories.

Note that we converted to MCQ or text generation some tasks, such as Part-of-Speech (POS) tagging and Named Entity Recognition (NER), which have been traditionally addressed as sequence labeling tasks. The aim was to ease evaluation as we currently cannot handle sequence labeling tasks (we plan support for this in the future). 

\paragraph{Reusing Public Datasets}
Some of the datasets are subsampled from publicly available test sets with preexisting prompts and ground-truth answers. This includes datasets from the Arabic subset of the xP3 dataset~\cite{muennighoff2022crosslingual}, from which we subsampled 68 datasets, covering 12 tasks, to include 25 development and 50 test examples. We further reformatted AraMath~\cite{alghamdi-liang-zhang:2022:LREC} to MCQ format, as an additional dataset.

\paragraph{Prompting Existing NLP Datasets}
We created natural language prompts based on publicly available Arabic NLP datasets using the PromptSource tool~\cite{bach2022promptsource}. 
We developed 2--8 different prompt templates per dataset, resulting in an equal number of sub-datasets. Figure~\ref{fig:prompt_examples} in the Appendix shows four different prompt templates developed for one of the datasets.

\paragraph{Translating English Datasets to Arabic}
Some of our datasets were created by translating existing English datasets to Arabic. We have a total of 483 such datasets, covering 29 different tasks, sampled from PromptSource \cite{bach2022promptsource}, Super-NaturalInstuctions \cite{supernaturalinstructions,naturalinstructions}, and TruthfulQA~\cite{lin2021truthfulqa}. The translations were evaluated both automatically and manually as described in \cite{el2024creating}. 

\paragraph{Developing New Datasets}
We further developed 16 brand new datasets with 1,755 prompts, covering specialized, structured, and rare examples to better test model generalization, e.g., to tasks such as grammatical error detection and factuality.

\paragraph{Augmenting with Synthetic Examples}
Our target was to have 10--100 test examples per dataset. However, for 14 datasets, we had less than 10 examples; we thus used GPT-4o to generate synthetic examples, which we checked manually.

\subsection{Quality Assurance}

To ensure data quality, we conducted extensive quality checks in three iteratively repeated stages:

\begin{itemize}[leftmargin=*]
\item \emph{Completeness:} We ensured that all required fields in all datasets were fully populated, with no missing or null values. We found that 1\% of our test examples contained null values, which we removed; we further found that 7\% of the datasets included duplicates, which we also removed.

\item \emph{Consistency:} We established a standardized format to maintain consistency across the datasets. We found that approximately 17\% of the datasets exhibited format-related issues, such as improper structure, or incorrect labels, 
which we fixed.

\item \emph{Reliability} We asked 16 annotators to conduct a manual review of random samples from each dataset checking that each \emph{instruction}, \emph{input}, and \emph{output} were clear and cultural appropriate. We found issues for 10\% of the datasets; to fix them, we edited some specific examples or excluded entire datasets.
\end{itemize}

\subsection{Mitigating Data Leakage}
A primary goal of the \balsam initiative is to establish a fair, unbiased, and trusted benchmark for evaluating LLMs in Arabic. Thus, it is critical to prevent test set leakage and to minimize the risk of contamination of LLM training data.

In order to protect the integrity and reliability of the benchmark, we restricted the access to the test sets to a small group of individuals responsible for quality assessment and platform development: in fact, the vast majority of members of the \balsam{} team only know the part of the raw test data candidates they contributed initially, but they have no access to the final test data. 

\section{Evaluation Setup}


\subsection{Benchmarking Phases}

The \balsam benchmark comprises a total of 37,419 test and 15,742 development examples and runs in two phases:

\begin{itemize}[leftmargin=*]
    \item \emph{Phase 1.}
    This phase includes 54 tasks across 13 categories focusing on text generation. It contains 13,121 test and 6,434 dev examples. The largest categories are \emph{creative writing} and \emph{translation}, which cover tasks such as \emph{story composition} and \emph{dialect translation}, respectively. A complete breakdown of the categories and associated tasks in this phase is given in Table~\ref{tab:phase1_dataset_statistics} in the Appendix.

    \item \emph{Phase 2:}
    This phase includes 50 tasks across 13 categories and contains 24,298 test examples and 9,308 development examples. The focus of this phase is on multiple-choice question answering and specific generation tasks(Diacritization,Translation/Transliteration). 
\end{itemize}

The two phases share 12 categories in common, with the remaining categories being \textit{translation} (unique to Phase~1) and \textit{factuality} (unique to Phase~2). A complete breakdown of all categories and tasks is provided in Table~\ref{tab:phase2_dataset_statistics} in the Appendix.

\begin{table*}[t!]
\centering
\small
\setlength{\tabcolsep}{4pt} 
\renewcommand{\arraystretch}{0.96}

\scalebox{0.9}{%
\begin{tabular}{lccccccccccccccc}
\toprule
\textbf{Model} &
\textbf{CW} & \textbf{ENT} & \textbf{FIB} & \textbf{IE} & \textbf{LOG} & \textbf{PE} & \textbf{QA} & \textbf{RC} & \textbf{ST} & \textbf{SUM} & \textbf{TC} & \textbf{TM} & \textbf{MT/TL} & \textbf{AVG} & \textbf{AVG*} \\
\midrule

SILMA-9B Instruct-v1.0  & 0.23 & 0.13 & 0.12 & 0.32 & \bf 0.22 & 0.66 & \bf 0.31 & \bf 0.55 & 0.20 & 0.20 &  0.36 & 0.60 & 0.13 & 0.31 & 0.33 \\
Nuha v2   &  0.22  & 0.12 & 0.12 &  0.32  & 0.20 &  0.81  & 0.25 & 0.35 & 0.28 & 0.19 & \bf 0.39 & \bf 0.64 &  0.15  & 0.31   & 0.32 \\
Jais-family 13B-chat    & \bf 0.24 & 0.08 & 0.10 & 0.25 & 0.17 & \bf 0.89 & 0.22 & 0.51 & 0.19 & \bf 0.26 & 0.15 & 0.48 &  --  &  --  & 0.30 \\
Command R+              & 0.19 & 0.07 & 0.11 & \bf 0.33 & 0.17 & 0.68 & \bf 0.31 & 0.41 & 0.28 & 0.16 & 0.28 & 0.53 & 0.15 & 0.28 & 0.29 \\
GPT-4o                  & 0.22 & 0.10 & 0.20 & 0.28 & 0.16 & 0.21 & 0.23 & 0.29 & \bf 0.38 & 0.17 & 0.30 &  0.62 & \bf 0.17 & 0.26 & 0.27 \\
Iron Horse GV V5a 
& 0.20 & 0.10 & \bf 0.21 & 0.27 & 0.15 & 0.48 & 0.21 & 0.24 & 0.36 & 0.15 & 0.27 & 0.56 & 0.14 & 0.26 & 0.27 \\
Yehia 7B Preview        & 0.23 & 0.13 & 0.18 & 0.26 & 0.20 & 0.34 & 0.23 & 0.28 & 0.26 & 0.20 & 0.24 &  0.62 & 0.14 & 0.25 & 0.27 \\
AceGPT-v2 8B Chat       & 0.19 & 0.11 & 0.14 & 0.29 & 0.18 & 0.49 & 0.25 & 0.36 & 0.19 & 0.19 & 0.23 & 0.51 & 0.11 & 0.25 & 0.26 \\
Grok-2-latest           & 0.20 & 0.08 & 0.14 & 0.23 & 0.15 & 0.16 & 0.22 & 0.29 & 0.30 & 0.18 & 0.18 & 0.49 & 0.14 & 0.21 & 0.24 \\
Gemini 2.0 Flash        & 0.17 & 0.06 & 0.14 & 0.28 & 0.15 & 0.13 & 0.25 & 0.30 & 0.33 & 0.15 & 0.24 & 0.33 & 0.13 & 0.20 & 0.22 \\
Mistral-saba-latest     & 0.21 & 0.07 & 0.16 & 0.18 & 0.14 & 0.15 & 0.20 & 0.23 & 0.29 & 0.18 & 0.19 & 0.55 & 0.15 & 0.21 & 0.21 \\
Claude Sonnet 3.5       & 0.13 & \bf 0.15 & 0.07 & 0.19 & 0.09 & 0.24 & 0.18 & 0.20 & 0.35 & 0.15 & 0.12 & 0.38 & 0.12 & 0.18 & 0.19 \\
Command-r7b 12-2024     & 0.17 & 0.07 & 0.15 & 0.15 & 0.13 & 0.26 & 0.15 & 0.22 & 0.19 & 0.16 & 0.13 & 0.41 & 0.13 & 0.18 & 0.19 \\
Gemma 2 9B              & 0.16 & 0.09 & 0.11 & 0.19 & 0.14 & 0.31 & 0.18 & 0.23 & 0.19 & 0.15 & 0.10 & 0.30 & 0.05 & 0.17 & 0.19 \\
Qwen 2.5 32B            & 0.14 & 0.09 & 0.13 & 0.16 & 0.11 & 0.30 & 0.15 & 0.13 & 0.23 & 0.16 & 0.08 & 0.43 & 0.08 & 0.17 & 0.18 \\
DeepSeek V3             & 0.17 & 0.12 & 0.11 & 0.18 & 0.11 & 0.12 & 0.15 & 0.14 & 0.25 & 0.15 & 0.08 & 0.40 & 0.15 & 0.16 & 0.17 \\
C4AI Aya Expanse 32B    & 0.14 & 0.07 & 0.11 & 0.13 & 0.07 & 0.23 & 0.14 & 0.25 & 0.13 & 0.19 & 0.06 & 0.38 & 0.10 & 0.15 & 0.16 \\
Fanar-C-1-8.7B          & 0.14 & 0.09 & 0.07 & 0.16 & 0.11 & 0.36 & 0.14 & 0.15 & 0.11 & 0.14 & 0.11 & 0.33 &  --  &  --  & 0.16 \\
Amazon Nova Pro         & 0.15 & 0.07 & 0.07 & 0.12 & 0.07 & 0.14 & 0.15 & 0.10 & 0.26 & 0.15 & 0.04 & 0.37 & 0.09 & 0.14 & 0.15 \\
Mistral Large       & 0.08 & 0.10 & 0.04 & 0.10 & 0.08 & 0.17 & 0.12 & 0.15 & 0.06 & 0.07 & 0.09 & 0.30 & 0.05 & 0.11 & 0.12 \\
DBRX-instruct           & 0.03 & 0.01 & 0.03 & 0.03 & 0.02 & 0.10 & 0.04 & 0.04 & 0.03 & 0.03 & 0.02 & 0.12 & 0.02 & 0.04 & 0.04 \\
Aragpt2 mega            &  --  & 0.11 & 0.04 &  --  & 0.04 &  --  & 0.05 & 0.06 & 0.04 & 0.13 & 0.06 & 0.33 &  --  &  --  & -- \\

\bottomrule
\end{tabular}
}
\caption{\textbf{Automatic evaluation across categories.} 
``--'' indicates that the model exceeded the token limits and did not complete the category.
List of categories: CW (Creative Writing), ENT (Entailment), FIB (Fill in the Blank), IE (Information Extraction), LOG (Logic), PE (Program Execution), QA (Question Answering), RC (Reading Comprehension), ST (Sequence Tagging), SUM (Summarization), TC (Text Classification), TM (Text Manipulation), MT/TL (Machine Translation/Transliteration), AVG (Average), AVG* (Average w/o Translation).
}
\label{tab:Models_results}
\end{table*}

\subsection{Evaluation Framework}

We adopted the LM-Evaluation-Harness \citep{eval-harness} framework, henceforth \textit{LM-Harness}, for several reasons: (\emph{i})~it supports evaluation of both open-source LLMs with accessible weights as well as commercial LLMs that are only available via API calls, (\emph{ii})~it allows flexible customization of tasks and benchmarks through YAML files, and (\emph{iii})~it has been used in various leaderboards on Hugging Face and as part of various LLM development pipelines, e.g.,~by Fanar \cite{fanar2024}.
 
\subsection{Evaluation Platform}

We enhanced the schema of \textit{LM-Harness}
\footnote{\url{https://github.com/ksaa-nlp/balsam-eval}} 
to standardize the input data. Each dataset file is assigned a unique ID, and its JSON content is preprocessed into the YAML format required by \textit{LM-Harness}, which includes task metadata and dataset split paths. The evaluation jobs on the platform are organized into categories, tasks, and datasets. Categories group related tasks for visualization purposes. Tasks represent specific objectives such as summarization, sequence tagging, title generation, and transliteration, while datasets contain data split by prompts and data items for each task. 

Users register models via an OpenAI-compatible API (requiring model ID and URL) or a public model (e.g., from \href{https://platform.aixplain.com/}{aiXplain})
with optional metadata such as model name and training data. Evaluation requests are run in parallel for selected categories to minimize waiting times. Results are calculated as task-level macro-averages of dataset scores. Similarly, category-level results are computed as the macro-average of per-task scores.  The overall score of a model is the macro-average score across all tasks. The \balsam Leaderboard
\footnote{\url{https://benchmarks.ksaa.gov.sa/b/balsam}} 
summarizes the model performance, displaying average scores for all tasks. Scores, ranging from 0 to 1, reflect task-specific metrics and enable clear comparisons of model performance across tasks.

\subsection{Evaluation Measures}

Given that the focus of Phase 1 on text generation, we began evaluation using BLEU \cite{papineni-etal-2002-bleu} for the translation category and ROUGE-LSum for the rest of categories \cite{lin-2004-rouge}. 
For analysis purposes, we also perform manual judgments (see below).

\section{Experiments}


\subsection{Experimental Setup}

We selected a comprehensive set of LLMs that support Arabic; see Appendix~\ref{sec:app_models_details} for a detailed list and description of the models we used.

\begin{itemize}[leftmargin=*]
    \item \emph{Open-weights models:} we chose them based on public availability, relevance to Arabic NLP, and architectural diversity. 
    We conducted all experiments using four NVIDIA A100 GPUs, each with 40G of VRAM.
    \item \emph{Closed models:} we included some popular ones that support Arabic, and we accessed them via their standard APIs or by provider request.
\end{itemize}

\begin{table*}[t]
\centering
\small
\renewcommand{\arraystretch}{0.96}

\setlength{\tabcolsep}{4pt} 
\scalebox{0.90}{%
\begin{tabular}{lcccccccccccccccc}
\toprule
\textbf{Model} &
\textbf{CW} & \textbf{ENT} & \textbf{FIB} & \textbf{IE} & \textbf{LOG} & \textbf{PE} & \textbf{QA} & \textbf{RC} & \textbf{ST} & \textbf{SUM} & \textbf{TC} & \textbf{TM} & \textbf{MT/TL} & \textbf{AVG} & \textbf{AVG*} \\
\midrule
GPT-4o & 2.78 & 3.00 & 2.50 & 2.57 & 2.50 & 2.30 & 2.30 & 2.65 & 2.12 & 2.72 & 2.57 & 2.72 & 2.75 & 2.58 & 2.56\\
Iron Horse GV V5a
& 2.63 & 3.00 & 2.50 & 2.32 & 2.15 & 2.50 & 2.25 & 2.77 & 2.08 & 2.52 & 2.23 & 2.52 & 2.85 & 2.49 & 2.46 \\
Claude Sonnet 3.50 & 2.83 & 2.52 & 2.57 & 2.58 & 2.27 & 2.58 & 2.35 & 2.53 & 2.02 & 2.62 & 1.97 & 2.67 & 2.68 & 2.48 & 2.46 \\
DeepSeek V3 & 2.70 & 2.93 & 2.17 & 2.57 & 2.20 & 2.30 & 2.37 & 2.80 & 1.97 & 2.88 & 1.87 & 2.52 & 2.48 & 2.44 & 2.44 \\
Nuha v2    &  2.75  & 2.62 & 2.53 &  2.38  & 1.95 &  2.20  & 2.32 & 2.83 & 2.02 & 2.88 & 1.70 & 2.70 &  2.63  & 2.42   & 2.40 \\
Grok-2-latest  & 2.78 & 3.00 & 1.95 & 2.47 & 2.52 & 2.50 & 2.27 & 2.80 & 1.80 & 2.75 & 1.60 & 2.47 & 2.53 & 2.42 & 2.41 \\
Gemini 2.0 Flash & 2.73 & 2.74 & 2.42 & 2.53 & 2.43 & 2.13 & 2.12 & 2.77 & 1.95 & 2.60 & 1.55 & 2.37 & 2.72 & 2.39 & 2.36 \\
Command R+ & 2.60 & 2.81 & 2.23 & 2.52 & 2.13 & 2.08 & 2.30 & 2.58 & 1.97 & 2.70 & 1.57 & 2.37 & 2.42 & 2.33 & 2.32 \\
Fanar-C-1-8.7B & 2.73 & 2.98 & 1.82 & 2.62 & 2.25 & 2.25 & 2.70 & 2.82 & 1.22 & 2.67 & 1.62 & 2.03 & -- & -- & 2.31\\
c4ai-aya-expanse-32b & 2.65 & 2.88 & 2.37 & 2.37 & 2.02 & 2.28 & 2.23 & 2.57 & 1.68 & 2.70 & 1.62 & 2.47 & 2.13 & 2.31 & 2.32 \\
Mistral-saba-latest & 2.60 & 2.86 & 2.15 & 2.55 & 2.00 & 1.25 & 2.38 & 2.82 & 1.90 & 2.78 & 1.43 & 2.53 & 2.50 & 2.29 & 2.27\\
Yehia-7B preview & 2.68 & 2.98 & 1.88 & 2.28 & 2.08 & 1.83 & 2.28 & 2.63 & 1.75 & 2.50 & 1.65 & 2.68 & 2.13 & 2.26 & 2.27 \\
Amazon Nova Pro & 2.65 & 2.86 & 2.23 & 2.20 & 2.18 & 1.42 & 2.32 & 2.63 & 1.78 & 2.75 & 1.60 & 2.35 & 2.42 & 2.26 & 2.25\\
Gemma2 9B & 2.62 & 2.90 & 1.70 & 2.33 & 2.08 & 1.97 & 2.20 & 2.85 & 1.73 & 2.67 & 1.77 & 1.93 & 2.05 & 2.22 & 2.23\\
Qwen-2.5 32b & 2.83 & 2.55 & 1.97 & 2.15 & 1.97 & 2.18 & 2.12 & 2.72 & 1.77 & 2.55 & 1.45 & 2.42 & 2.08 & 2.21 & 2.22 \\
Command-r7b 12-2024 & 2.62 & 2.83 & 1.60 & 2.08 & 1.88 & 2.00 & 2.20 & 2.45 & 1.75 & 2.77 & 1.18 & 2.38 & 1.87 & 2.12 & 2.15\\
Jais-family 13b-chat & 2.03 & 2.88 & 1.13 & 2.23 & 1.70 & 2.17 & 1.87 & 2.52 & 1.35 & 2.38 & 1.02 & 2.18 & -- & -- & 1.96\\
SILMA-9B Instruct-v1.0  & 2.33 & 2.00 & 1.42 & 2.1 & 1.73 & 1.68 & 1.83 & 2.13 & 1.52 & 2.4 & 1.63 & 2.28 & 2.00 & 1.93 & 1.92 \\
AceGPT-v2-8B-Chat & 2.17 & 2.21 & 1.08 & 2.17 & 1.75 & 1.38 & 1.50 & 2.57 & 1.63 & 2.62 & 1.07 & 2.05 & 1.77 & 1.84 & 1.85 \\
Mistral large & 1.20 & 1.79 & 0.80 & 0.98 & 1.22 & 0.98 & 1.65 & 1.52 & 0.65 & 0.58 & 0.62 & 1.27 & 1.78 & 1.16 & 1.11 \\
DBRX-instruct & 0.23 & 0.24 & 0.07 & 0.22 & 0.28 & 0.73 & 0.43 & 0.18 & 0.77 & 0 & 0.22 & 0.12 & 1.28 & 0.37 & 0.29\\
Aragpt2-mega & -- & 0.14 & 0.13 & -- & 0.13 & -- & 0.13 & 0.42 & 0.1 & 1.63 & 0.05 & 0.37 & -- & -- & -- \\

\bottomrule
\end{tabular}
}
\caption{\textbf{Manual evaluation (3 evaluators; 20 examples per category).}
``--'' indicates that the model exceeded token limits and did not complete the category.
List of categories: CW (Creative Writing), ENT (Entailment), FIB (Fill in the Blank), IE (Information Extraction), LOG (Logic), PE (Program Execution), QA (Question Answering), RC (Reading Comprehension), ST (Sequence Tagging), SUM (Summarization), TC (Text Classification), TM (Text Manipulation), MT/TL (Machine Translation/Transliteration), AVG (Average), AVG* (Average w/o Translation).
}
\label{tab:Model-results-human-judgement}
\end{table*}
\vspace{-6pt}
\subsection{Results and Discussion}

\paragraph{Challenges in Automatic Evaluation.}
Table~\ref{tab:Models_results} shows the automatic evaluation results of the LLMs across 13 categories using ROUGE-LSum and BLEU. Unexpectedly, the results show that SILMA-9B is far ahead of much larger models such as Aya 32B, Qwen-2.5 32B, and DeepSeek V3. This prompted us to manually examine random output samples to better understand the underlying reasons. Our analysis revealed the following:
\begin{itemize}[leftmargin=*]
    \item SILMA-9B's output was generally terse, while the outputs of the other models were verbose; the metrics naturally preferred shorter answers. 
    In a Question Answering example where the correct answer was \<باريس> (Paris), SILMA-9B gave a matching terse reply, while other models provided more detailed, verbose answers with 25 words or longer (Full example iin Appendix \ref{appendix:example-outputs}).
    \item BLEU uses the geometric mean of unigram to 4-gram precisions. Because many gold answers were short, trigram and 4-gram matches were often absent, causing BLEU scores to be zero despite matching unigrams and bigrams.
    \item BLEU and ROUGE rely on exact word matches, which is difficult for Arabic's complex morphology. For example, the reference \<كتاب> (`book') and the prediction \<الكتاب> (`the book') do not match exactly.
\end{itemize}

\paragraph{Human Evaluation.}
Next, we conducted a manual evaluation on a random sample of the test set, composed of 20 questions per category, where humans would rate the outputs from all LLMs. The correctness of each output, on a 0--3 scale, was judged by three judges. Thus, the total number of performed judgments was 254 questions $\times$ 22 LLMs $\times$ 3 judges = 16,764 judgments. The detailed annotation instructions we gave to the judges are given in Appendix~\ref{app:anno_instruct}.

The average score per model from these judgments are reported in Table~\ref{tab:Model-results-human-judgement}, where we can see that GPT-4o achieves the highest average score. 

\paragraph{Human-to-Automatic Measure Correlation.}
We measured the Pearson correlation of human judgments against ROUGE-LSum and BLEU.
Table \ref{tab:correlation-between-judges} shows the correlation between the three human judgments across categories. The average correlation between the judges is 0.75, and they correlated more with each other for some categories compared to others. For example, Creative Writing had the lowest correlation (0.636), while Reading Comprehension had the highest correlation (0.88).
Table \ref{tab:correl-rouge-bleu} lists the correlations of manual evaluation against ROUGE-LSum and BLEU.  Since we had three judges, we computed the correlation between the metrics and the average judges' scores. We can see very poor correlation between manual judgments and automatic measures. 

\paragraph{Beyond BLEU and ROUGE.} 
We explored some alternative evaluation approaches, namely: 

\begin{itemize}[leftmargin=*]
\item \textbf{Semantic Evaluation:} We used BERTScore~\cite{zhang2020bertscore}, which captures semantic similarity more effectively than surface-level $n$-gram overlap.
\item \textbf{LLM-Based Answer Extraction:} We used Gemini 2.5 Flash (zero-shot, no chain-of-thought) to extract concise answers from the model-generated outputs. We used the prompt reported in the Appendix, Listing \ref{lst:prompt_for_answer}.  

\item \textbf{LLM-Based Scoring:} We used Gemini 2.5 Flash to rate the extracted answers on a 0–3 scale, mirroring the manual evaluation scheme.\footnote{We also experimented with GPT-4o and GPT-4o mini as LLM judges. GPT-4 and Gemini showed nearly identical correlation with human scores, both outperforming GPT-4o mini by a sizable margin. Eventually, we selected Gemini 2.5 Flash due to its substantially lower cost.} The scoring prompt is shown in Appendix Listing \ref{box:llm-as-a-judge}.
\end{itemize}

\begin{table}[t]
    \centering
    \small
    \renewcommand{\arraystretch}{0.96}

    \begin{tabular}{lcccc}
    \toprule
 \textbf{Category} & \textbf{1 \& 2} & \textbf{1 \& 3} & \textbf{2 \& 3} & \textbf{Avg.} \\ \midrule
Creative Writing & 0.579 & 0.563 & 0.765 & 0.636\\
Entailment & 0.824 & 0.757 & 0.768 & 0.783\\
Fill in the Blank & 0.587 & 0.659 & 0.826 & 0.691\\
Info. Extraction & 0.636 & 0.602 & 0.730 & 0.656\\
Logic & 0.578 & 0.630 & 0.586 & 0.598\\
Program Execution & 0.722 & 0.697 & 0.841 & 0.753\\
Q\&A & 0.883 & 0.813 & 0.816 & 0.837\\
Reading Compr. & 0.885 & 0.894 & 0.860 & 0.880\\
Sequence Tagging & 0.738 & 0.768 & 0.935 & 0.814\\
Summarization & 0.828 & 0.774 & 0.754 & 0.785\\
Text Classification & 0.833 & 0.820 & 0.921 & 0.858\\
Text Manipulation & 0.790 & 0.808 & 0.792 & 0.797\\
Translation & 0.646 & 0.607 & 0.746 & 0.666\\ \midrule
\textbf{Average} & 0.733 & 0.722 & 0.795 & 0.750  \\ \bottomrule
    \end{tabular}
    \caption{Correlation between the three human judges (1,~2,~\& 3) per category.} 
    \label{tab:correlation-between-judges}
\end{table}

\begin{table}[t]
    \centering
    \small
    \renewcommand{\arraystretch}{0.96}

    \begin{tabular}{lcc}
    \toprule
\textbf{Category}	&	\textbf{ROUGE-Lsum}	&	\textbf{BLEU}	\\ \midrule
Creative Writing	&	-0.509	&	-0.613	\\
Entailment	&	-0.300	&	0.010	\\
Fill in the Blank	&	-0.033	&	-0.008	\\
Info. Extraction	&	0.139	&	0.514	\\
Logic	&	0.425	&	0.296	\\
Program Execution	&	-0.151	&	-0.005	\\
Question Answering	&	0.339	&	0.316	\\
Reading Comprehension	&	0.318	&	0.299	\\
Sequence Tagging	&	0.537	&	0.094	\\
Summarization	&	-0.393	&	-0.187	\\
Text Classification	&	0.100	&	0.090	\\
Text Manipulation	&	0.462	&	0.460	\\
Translation	&	0.506	&	0.481	\\ \midrule
Average	&	0.111	&	0.134	\\
\bottomrule
    \end{tabular}
    \caption{Correlation of human judgments against ROUGE-LSum and BLEU for different categories.}
    \label{tab:correl-rouge-bleu}
\end{table}

\begin{table*}[t]
    \centering
    \small
    \renewcommand{\arraystretch}{0.9}

    \begin{tabular}{lrrrrrrc}
    \toprule
\textbf{Category}	&	
\textbf{ROUGE} & \textbf{Ext. ROUGE} & \textbf{BLEU} & \textbf{Ext. BLEU} & \textbf{BERT} & \textbf{Ext. BERT} & \textbf{LLM-J} \\
\midrule
Creative Writing	&	-0.509	&	-0.476	&	-0.613	&	-0.582	&	-0.629	&	-0.392	&	\textbf{0.824}	\\
Entailment	&	-0.300	&	0.227	&	0.010	&	0.546	&	-0.244	&	-0.176	&	\textbf{0.950}	\\
Fill in the Blank	&	-0.033	&	0.390	&	-0.008	&	-0.502	&	0.386	&	0.696	&	\textbf{0.944}	\\
Information Extraction	&	0.139	&	0.656	&	0.514	&	0.766	&	0.034	&	0.691	&	\textbf{0.824}	\\
Logic	&	0.425	&	0.742	&	0.296	&	0.554	&	0.429	&	0.676	&	\textbf{0.945}	\\
Program Execution	&	-0.151	&	0.715	&	-0.005	&	0.882	&	-0.235	&	-0.034	&	\textbf{0.911}	\\
Question Answering	&	0.339	&	0.807	&	0.316	&	0.494	&	0.408	&	0.852	&	\textbf{0.977}	\\
Reading Comprehension	&	0.318	&	0.285	&	0.299	&	0.008	&	0.413	&	0.268	&	\textbf{0.931}	\\
Sequence Tagging	&	0.537	&	-0.241	&	0.094	&	-0.793	&	0.691	&	0.182	&	\textbf{0.931}	\\
Summarization	&	-0.393	&	-0.754	&	-0.187	&	-0.676	&	0.092	&	-0.604	&	\textbf{0.934}	\\
Text Classification	&	0.100	&	0.400	&	0.090	&	0.275	&	0.251	&	0.830	&	\textbf{0.948}	\\
Text Manipulation	&	0.462	&	0.677	&	0.460	&	0.685	&	0.401	&	0.678	&	\textbf{0.919}	\\
Translation	&	0.506	&	0.806	&	0.481	&	0.754	&	0.390	&	0.831	&	\textbf{0.899}	\\ \midrule
Average	&	0.111	&	0.326	&	0.134	&	0.186	&	0.184	&	0.346	&	\textbf{0.918}	\\
\bottomrule
    \end{tabular}
    \caption{Correlation of human judgments against ROUGE-LSum, BLEU, BERTScore (and their extracted versions), and LLM-as-a-Judge (LLM-J).
    }
    \label{table:correlations-all}
\end{table*}

\begin{table*}[htbp!]
\centering
\renewcommand{\arraystretch}{0.96}

\small
\setlength{\tabcolsep}{4pt} 
\scalebox{0.90}{%
\begin{tabular}{lcccccccccccccccc}
\toprule
\textbf{Model} &
\textbf{CW} & \textbf{ENT} & \textbf{FIB} & \textbf{IE} & \textbf{LOG} & \textbf{PE} & \textbf{QA} & \textbf{RC} & \textbf{ST} & \textbf{SUM} & \textbf{TC} & \textbf{TM} & \textbf{MT/TL} & \textbf{AVG} & \textbf{AVG*} \\

\midrule
GPT-4o  & 1.93 & 2.14 & \textbf{1.77} & 2.14 & 1.92 & 1.81 & 2.16 & 2.21 & \textbf{1.99} & 1.98 & \textbf{2.23} & 2.02 & 2.3 & \textbf{2.05} & \textbf{2.03} \\

Gemini 2.0 Flash  & \textbf{1.96} & 2.00 & 1.55 & 2.15 & 1.91 & 2.18 & \textbf{2.20} & \textbf{2.27} & 1.85 & 1.99 & 2.03 & 1.98 & 2.24 & 2.02 & 2.01 \\
Iron Horse GV V5a
& 1.90 & 2.14 & 1.35 & \textbf{2.17} & 1.88 & \textbf{2.56} & 2.12 & 2.05 & 1.82 & 1.89 & 1.90 & 2.02 & \textbf{2.51} & 2.02 & 1.98 \\
DeepSeek V3  & 1.7 & 2.21 & 1.52 & 2.1 & 1.88 & 2.32 & 2.01 & 2.11 & 1.83 & 1.95 & 2.04 & 2.02 & 2.21 & 1.99 & 1.97 \\
Claude Sonnet 3.5  & 1.85 & 2.07 & 1.32 & 2.08 & 1.8 & 2.42 & 2.09 & 2.18 & 1.88 & 1.95 & 1.79 & \textbf{2.09} & 2.37 & 1.99 & 1.96 \\

Grok-2-latest  & 1.94 & 2.07 & 1.29 & 2.10 & \textbf{2.01} & 2.15 & 2.04 & 2.22 & 1.59 & 1.98 & 2.07 & 1.86 & 2.10 & 1.96 & 1.94 \\
Nuha v2   &  1.86  & 1.86 & 1.39 &  1.99  & 1.84 &  2.37  & 1.91 & 2.20 & 1.59 & 1.95 & \textbf{2.20} & 1.86 &  1.96  & 1.92   & 1.92 \\
Qwen-2.5 32b  & 1.85 & 1.93 & 1.39 & 1.88 & 1.82 & 1.88 & 1.79 & 2.02 & 1.57 & 1.96 & 1.77 & 1.78 & 1.74 & 1.8 & 1.8 \\
Mistral-saba-latest  & 1.82 & 1.93 & 1.39 & 1.98 & 1.68 & 1.43 & 1.98 & 2.12 & 1.6 & 1.84 & 1.95 & 1.84 & 2.06 & 1.81 & 1.79 \\
Gemma2 9B  & 1.78 & \textbf{2.29} & 1.26 & 1.94 & 1.67 & 1.61 & 1.72 & 2.15 & 1.41 & 1.96 & 1.72 & 1.62 & 1.67 & 1.75 & 1.76 \\
c4ai-aya-expanse-32b  & 1.71 & 1.93 & 1.03 & 1.90 & 1.58 & 2.01 & 1.8 & 1.99 & 1.20 & 2.02 & 1.64 & 1.87 & 2.14 & 1.75 & 1.72 \\
Command R+  & 1.76 & 1.79 & 0.94 & 1.96 & 1.54 & 1.7 & 1.85 & 2.03 & 1.41 & 1.74 & 1.57 & 1.82 & 2.35 & 1.73 & 1.68 \\
Amazon Nova Pro  & 1.77 & 2.07 & 1.13 & 1.81 & 1.54 & 1.35 & 1.81 & 1.81 & 1.49 & 1.65 & 1.68 & 1.95 & 2.18 & 1.71 & 1.67 \\
Yehia-7B preview  & 1.79 & 2.14 & 0.9 & 1.89 & 1.46 & 1.34 & 1.73 & 2.06 & 1.17 & 1.83 & 1.62 & 1.83 & 2.02 & 1.68 & 1.65 \\

Fanar-C-1-8.7B  & 1.70 & 1.93 & 0.90 & 1.88 & 1.53 & 1.96 & 1.72 & 1.79 & 0.95 & 1.86 & 1.52 & 1.71 &  -  &  -  & 1.62 \\
Jais-family 13b-chat  & 1.80 & 1.86 & 0.52 & 1.62 & 1.39 & 2.42 & 1.49 & 1.85 & 0.66 & \textbf{2.05} & 1.11 & 1.57 &  -  &  -  & 1.53 \\
SILMA-9B Instruct-v1.0  & 1.67 & 1.57 & 0.97 & 1.63 & 1.50 & 1.31 & 1.46 & 2.17 & 1.01 & 1.73 & 1.77 & 1.5 & 1.84 & 1.55 & 1.52 \\
Command-r7b 12-2024  & 1.57 & 1.79 & 0.65 & 1.62 & 1.45 & 1.56 & 1.64 & 1.94 & 1.05 & 1.58 & 1.15 & 1.67 & 2 & 1.51 & 1.47 \\
Mistral large   & 1.52 & 1.21 & 1.13 & 1.65 & 1.54 & 1.50 & 1.57 & 1.86 & 1.04 & 1.52 & 1.51 & 1.16 & 1.47 & 1.44 & 1.43 \\
AceGPT-v2-8B-Chat  & 1.56 & 1.71 & 0.58 & 1.73 & 1.27 & 0.92 & 1.62 & 1.85 & 0.88 & 1.74 & 0.95 & 1.54 & 1.72 & 1.39 & 1.36 \\
DBRX-instruct  & 0.73 & 0.93 & 0.33 & 1.10 & 0.81 & 0.96 & 1.09 & 1.4 & 0.85 & 0.78 & 1.18 & 0.67 & 1.14 & 0.92 & 0.90 \\
Aragpt2-mega  &  -  & 0.00 & 0.03 &  -  & 0.05 &  -  & 0.12 & 0.25 & 0.02 & 0.15 & 0.15 & 0.29 &  -  &  -  &  -  \\

\bottomrule
\end{tabular}
}
\caption{\textbf{LLM-as-a-judge evaluation.} 
 ``--'' indicates that the model exceeded token limits and did not complete the category.
List of categories: CW (Creative Writing), ENT (Entailment), FIB (Fill in the Blank), IE (Information Extraction), LOG (Logic), PE (Program Execution), QA (Question Answering), RC (Reading Comprehension), ST (Sequence Tagging), SUM (Summarization), TC (Text Classification), TM (Text Manipulation), MT/TL (Machine Translation/Transliteration), AVG (Average), AVG* (Average w/o Translation).}
\label{tab:Model-results-LLM-as-a-judge}
\end{table*}

Table \ref{table:correlations-all} shows the correlation of human evaluation with ROUGE-LSum, BLEU, and BERTScore (with and without extraction of answers using an LLM) and LLM as a judge. We make the following observations:
\begin{itemize}[leftmargin=*]
    \item Using an LLM to extract the answer from the LLM output generally had a positive impact on correlation for all measures (ROUGE-LSum, BLEU, and BERTScore).
    \item BERTScore correlated better with human judgments compared to ROUGE and BLEU.
    \item The correlation for ROUGE-LSum, BLEU, and BERTScore varied widely from category to category, and the average was low.
    \item LLM as a judge was highly correlated with human judgments for all categories, with values ranging between 0.824 and 0.977. In fact, it correlated better with the average of judges' scores than judges correlated with each other.
\end{itemize}

Based on the above, we decided to drop ROUGE, BLEU, and BERTScore and rely solely on LLM as a Judge to evaluate the  LLMs. 
Table \ref{tab:Model-results-LLM-as-a-judge} lists the results for all models on the entire \balsam{} test set using LLM as a judge.  When comparing the results of using ROUGE-LSum and BLEU (Table~\ref{tab:Models_results}) to using LLM as a judge (Table~\ref{tab:Model-results-LLM-as-a-judge}), we can see that the order of LLMs changes completely. In fact, the top performer in Table \ref{tab:Models_results}, namely SILMA-9B-IT came out in the lower third in Table~\ref{tab:Model-results-LLM-as-a-judge}. Given the aforementioned discussion, the LLM as a judge results are more trustworthy as they correlate much better with human judgments.  

The results show that large closed models such as GPT-4o, Gemini 2.0, and DeepSeek V3 outperform all smaller Arabic-centric models such as Jais and Fanar by sizable margins.  Two large models, namely Mistral large  and DBRX-instruct (132B) performed poorly, trailing most other models.  This suggests that the model size is not a sufficient predictor of performance.  Some of the most likely factors that come into play are Arabic tokenization, size of Arabic training set, and Arabic-centric supervised fine-tuning.  

The results show some variability of how models generally perform for certain categories compared to others.  For example, models overall perform better on some tasks, such as \emph{translation} and \emph{entailment}, and worse on others, such as \emph{fill in the blank}.  Some models are relatively more capable for some categories compared to others.  For example, Grok-2 leads the pack for Logic and Iron Horse leads for Program Execution.  Similarly, some models rank higher for some categories and much lower in others.  For example, Jais and Fanar performed well for Summarization but poorly for Sequence Tagging. Some models performed poorly across the board, such as Aragpt2-mega and DBRX-Instruct.

\section{Conclusion and Future Directions }

We have presented \balsam --- a major collaborative effort to establish benchmarking standards and foster unity in LLM development and evaluation for Arabic. \balsam marks a significant step forward, offering evaluation across 78 tasks from 14 categories, with 37K development and 15K test examples.
It further offers an integrated platform, and Arabic LLM Leaderboard that enable effective evaluation, comparison, and progress tracking with reliable LLM-as-a-judge based evaluation.
However, challenges remain in enhancing data quality, addressing Arabic's linguistic diversity, and expanding the scope of tasks covered. 

In future work, we aim to improve dataset quality (e.g., eliminate translations and any form of synthetic data generation)  to add additional tasks, as well as to address the limitations listed in the next section.

\newpage
\section*{Limitations}

Our study provides insights into LLM performance; however, several key limitations warrant consideration and will be the focus of the next iteration of the \balsam benchmarking test sets. 
\begin{itemize}[leftmargin=*]
    \item Token length restrictions in certain models precluded their complete participation across all evaluation tasks, particularly affecting models with restricted context windows and preventing calculation of comprehensive performance scores for these systems.
    \item While efforts have been made to ensure the accuracy and neutrality of the datasets, we acknowledge the potential for unintended biases, particularly those arising from translated datasets that may have translation errors or cultural misalignments. For example, certain phrases, such as ``the Messenger of Islam Muhammad'' were identified as potentially problematic, as they may not align with widely accepted terminologies within specific cultural and religious contexts, such as the more commonly used ``Prophet Muhammad'' in Arabic and Islamic discourse. 
    \item Though BALSAM benchmarks LLMs across a variety of categories, some notable other functions and features of LLMs need to be considered such as fluency of the generated output, cultural alignment, ability to answer religious questions, ability to chat in a multi-turn scenario, propensity to hallucinate, tool usage, structured output generation, and many others.  We plan to address many of these aspects in the next iteration of BALSAM with new test sets.
\end{itemize}

\bibliography{custom,anthology}

\clearpage
\newpage
\appendix

\begin{table}[t]
\vspace{-0.2cm}
\centering
\scalebox{0.5}{ 
\begin{tabular}{lllrr}
\toprule
\textbf{No.} & \textbf{Category} & \textbf{Task} & \textbf{Test} & \textbf{Dev} \\ \midrule
\multirow{17}{*}{1} & \multirow{17}{*}{Creative Writing} & Definition Generation & 22 & 22 \\
& & Dialogue Generation & 146 & 65 \\
& & Explanation & 64 & 21 \\
& & Instruction Generation & 10 & 4 \\
& & Misc. & 21 & 9 \\
& & News Article Generation & 12 & 12 \\
& & Poem Generation & 25 & 9 \\
& & Question Generation & 1146 & 483 \\
& & Question Rewriting & 48 & 20 \\
& & Sentence Composition & 235 & 94 \\
& & Sentence Compression & 21 & 10 \\
& & Story Composition & 430 & 207 \\
& & Subject Generation & 497 & 232 \\
& & Text Completion & 119 & 46 \\
& & Text Generation & 130 & 92 \\
& & Wrong Candidate Generation & 233 & 93 \\
\midrule
\multirow{1}{*}{2} & \multirow{1}{*}{Entailment} & Textual Entailment & 14 & 13 \\
\midrule
\multirow{1}{*}{3} & \multirow{1}{*}{Fill in the Blank} & Fill in The Blank & 31 & 10 \\
\midrule
\multirow{6}{*}{4} & \multirow{6}{*}{Information Extraction} & Coreference Resolution & 18 & 7 \\
& & Disease Mention Identification & 10 & 9 \\
& & Keyword Extraction & 47 & 43 \\
& & Named Entity Recognition & 161 & 74 \\
& & Question Understanding & 22 & 10 \\
& & Relation Extraction & 10 & 9 \\
& & Extracting Required Information & 335 & 146 \\

\midrule
\multirow{6}{*}{5} & \multirow{6}{*}{Logic} & Cause Effect Classification & 39 & 18 \\
& & Coreference Resolution & 13 & 6 \\
& & Misc. & 69 & 29 \\
& & Predictive Analysis & 10 & 10 \\
& & Riddle Solving & 48 & 25 \\
& & Sentence Ordering & 18 & 8 \\
\midrule
\multirow{3}{*}{6} & \multirow{3}{*}{Translation/Transliteration} & Dialect Translation & 1200 & 600 \\
& & Machine Translation & 1810 & 646 \\
& & Transliteration & 220 & 220 \\
\midrule
\multirow{1}{*}{7} & \multirow{1}{*}{Program Execution} & Program Execution & 646 & 268 \\ 
\midrule
\multirow{2}{*}{8} & \multirow{2}{*}{Question Answering} & Answering Given Question & 2600 & 1484 \\
& & Question Decomposition & 10 & 2 \\
\midrule
\multirow{1}{*}{9} & \multirow{1}{*}{Reading Comprehension} & Reading Comprehension & 492 & 218 \\
\midrule
\multirow{2}{*}{10} & \multirow{2}{*}{Sequence Tagging} & Grammar Detection & 277 & 129 \\
& & Keyword Extraction & 58 & 20 \\
\midrule
\multirow{5}{*}{11} & \multirow{5}{*}{Summarization} & Text Summarization & 618 & 399 \\
& & Answer Extraction & 10 & 5 \\
& & Subject Generation & 10 & 3 \\
& & Subject Identification & 10 & 8 \\
& & Topic Identification & 23 & 18 \\
\midrule
\multirow{9}{*}{12} & \multirow{9}{*}{Text Classification} & Command Interpretation & 23 & 23 \\
& & Dialect Identification & 27 & 27 \\
& & Emotion Detection & 10 & 9 \\
& & Intent Classification & 10 & 4 \\
& & Offensive Language Detection & 21 & 11 \\
& & Problem Identification & 10 & 8 \\
& & Sarcasm Detection & 17 & 12 \\
& & Sentiment Analysis & 10 & 2 \\
& & Text Categorization & 56 & 23 \\
\midrule
\multirow{6}{*}{13} & \multirow{6}{*}{Text Manipulation} & Gender Rewriting & 347 & 119 \\
& & Grammar Correction & 269 & 202 \\
& & Intent Classification & 18 & 5 \\
& & Paraphrasing & 117 & 58 \\
& & Question Rewriting & 100 & 34 \\
& & Text Simplification & 98 & 41 \\
\bottomrule
\multicolumn{3}{c}{\text{Total}} & \text{13,121} & \text{6,434} \\ \hline
\end{tabular}
}
\caption{BALSAM Phase 1 benchmark dataset statistics}
\label{tab:phase1_dataset_statistics}
\end{table}

\begin{table}[t]
\vspace{-0.2cm}
\centering
\scalebox{0.5}{
\begin{tabular}{lllrr}
\toprule
\textbf{No.} & \textbf{Category} & \textbf{Task} & \textbf{Test} & \textbf{Dev} \\
\midrule
\multirow{5}{*}{1} & \multirow{5}{*}{Creative Writing} & Dialogue Generation & 72 & 30 \\
 & & Explanation & 25 & 10 \\
 & & Text Completion & 50 & 20 \\
 & & Text Continuation Evaluation & 10 & 10 \\
\midrule
\multirow{3}{*}{2} & \multirow{3}{*}{Entailment} & Duplicate Question Identification & 20 & 20 \\
 & & Semantic Similarity & 150 & 150 \\
 & & Textual Entailment & 305 & 150 \\
\midrule
\multirow{4}{*}{3} & \multirow{4}{*}{Factuality} & Answer Verification  & 50 & 20 \\
 & & Answerability Classification  & 25 & 10 \\
 & & Claim Verification  & 170 & 95 \\
 & & Text Classification  & 100 & 49 \\
\midrule
\multirow{1}{*}{3} & \multirow{1}{*}{Fill in the Blank} & Fill in The Blank & 25 & 10 \\
 & & Discourse Connective Identification  & 10 & 4 \\
\midrule

\multirow{2}{*}{4} & \multirow{2}{*}{Information Extraction} & Disease Mention Identification & 10 & 10 \\
 & & Named Entity Recognition & 10 & 10 \\
 & & Entity Categorization  & 10 & 10 \\
 & & Entity Recognition and Gender Identification  & 30 & 30 \\
 & & Entity Relation Classification & 25 & 10 \\
 & & Extracting Required Information  & 35 & 20 \\
  & & Text Classification  & 188 & 44 
\\ \midrule
\multirow{1}{*}{5} & \multirow{1}{*}{Logic} & Cause Effect Classification & 350 & 175 \\
& & Coherence Classification  & 50 & 20 \\
 & & Commonsense Validation  & 130 & 80 \\
  & & Evidence Evaluation  & 50 & 25 \\
 & & Logical Reasoning  & 30 & 30 \\
  & & Natural Language Inference  & 35 & 35 \\
\midrule
\multirow{1}{*}{6} & \multirow{1}{*}{Translation/Transliteration} & Machine Translation & 12890 & 3225 \\
\midrule
\multirow{1}{*}{7} & \multirow{1}{*}{Program Execution} & Program Execution & 25 & 10 \\
\midrule
\multirow{1}{*}{8} & \multirow{1}{*}{Question Answering} & Answering Given Question & 4979 & 2117 \\
\midrule
\multirow{4}{*}{9} & \multirow{4}{*}{Reading Comprehension}& Answer Verification & 25 & 10 \\
 & & Answerability Classification  & 75 & 30 \\
 & & Question Understanding  & 25 & 10 \\
&& Reading Comprehension & 350 & 250 \\
\midrule
\multirow{1}{*}{10} & \multirow{1}{*}{Sequence Tagging} & Sequence Tagging  & 100 & 25 
\\ \midrule

\multirow{15}{*}{10} & \multirow{15}{*}{Text Classification} & Dialect Identification & 490 & 228 \\
  & & Dialogue Act Recognition  & 25 & 10 \\
 & & Emotion Detection & 100 & 100 \\

 & & Ethics Classification  & 50 & 20 \\
 & & Hate Speech Detection  & 80 & 80 \\
 & & Offensive Language Detection & 200 & 110 \\
 & & Query Classification  & 50 & 24 \\
 & & Question Categorization  & 10 & 10 \\
 & & Question Understanding  & 25 & 10 \\
 & & Review Rating Prediction  & 30 & 30 \\
 & & Sarcasm Detection & 70 & 70 \\
 & & Sentiment Analysis & 605 & 509 \\
 & & Text Categorization & 235 & 110 \\
  & & Text Classification  & 1584 & 983 \\
 & & Topic Identification  & 10 & 10 
\\ 
\midrule
\multirow{1}{*}{13} & \multirow{1}{*}{Text Manipulation} & Diacritization  & 300 & 250
\\ \midrule
\multicolumn{3}{c}{\textbf{Total}} & \textbf{24298} & \textbf{9,308} \\
\bottomrule
\end{tabular}
}
\caption{BALSAM Phase 2 benchmark dataset statistics.}
\label{tab:phase2_dataset_statistics}
\end{table}

\section{Examples of Prompts}
\label{sec:app_prompt_example}
\begin{figure}[th]
    \centering
    \includegraphics[width=0.9\linewidth]{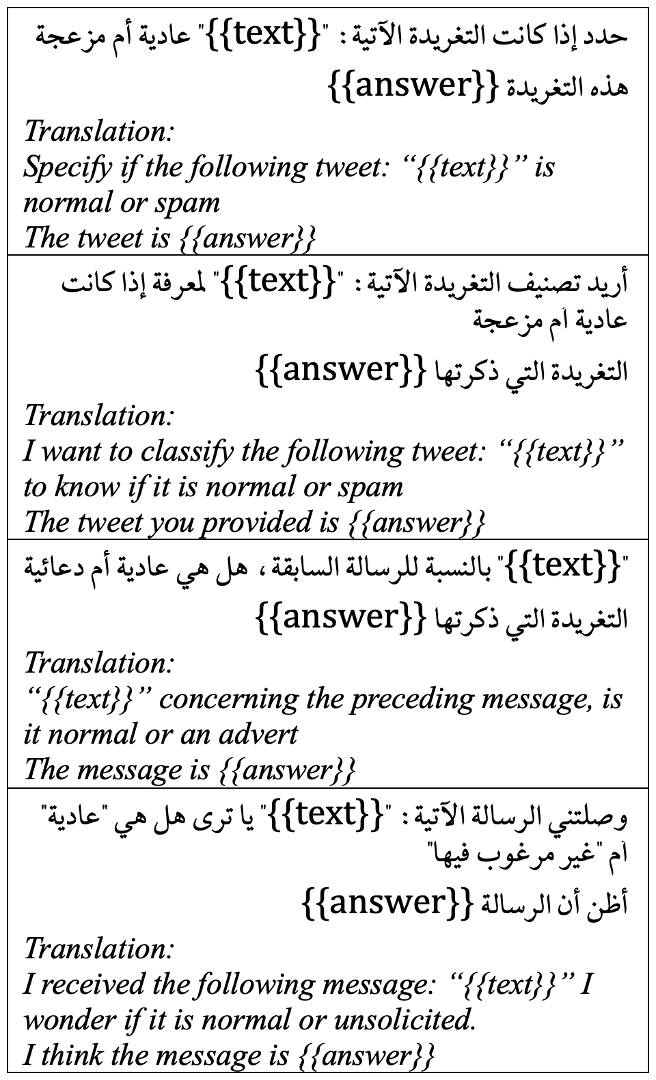}
    \caption{Example prompts for the Arabic tweet classification task.}
    \label{fig:prompt_examples}
\end{figure}

Figure \ref{fig:prompt_examples} shows some prompt templates that we used to create some of the datasets.

\section{Examples of Samples}
\label{sec:app_example_data_samples}
\begin{figure*}[t]
    \centering
    \includegraphics[width=0.8\linewidth]{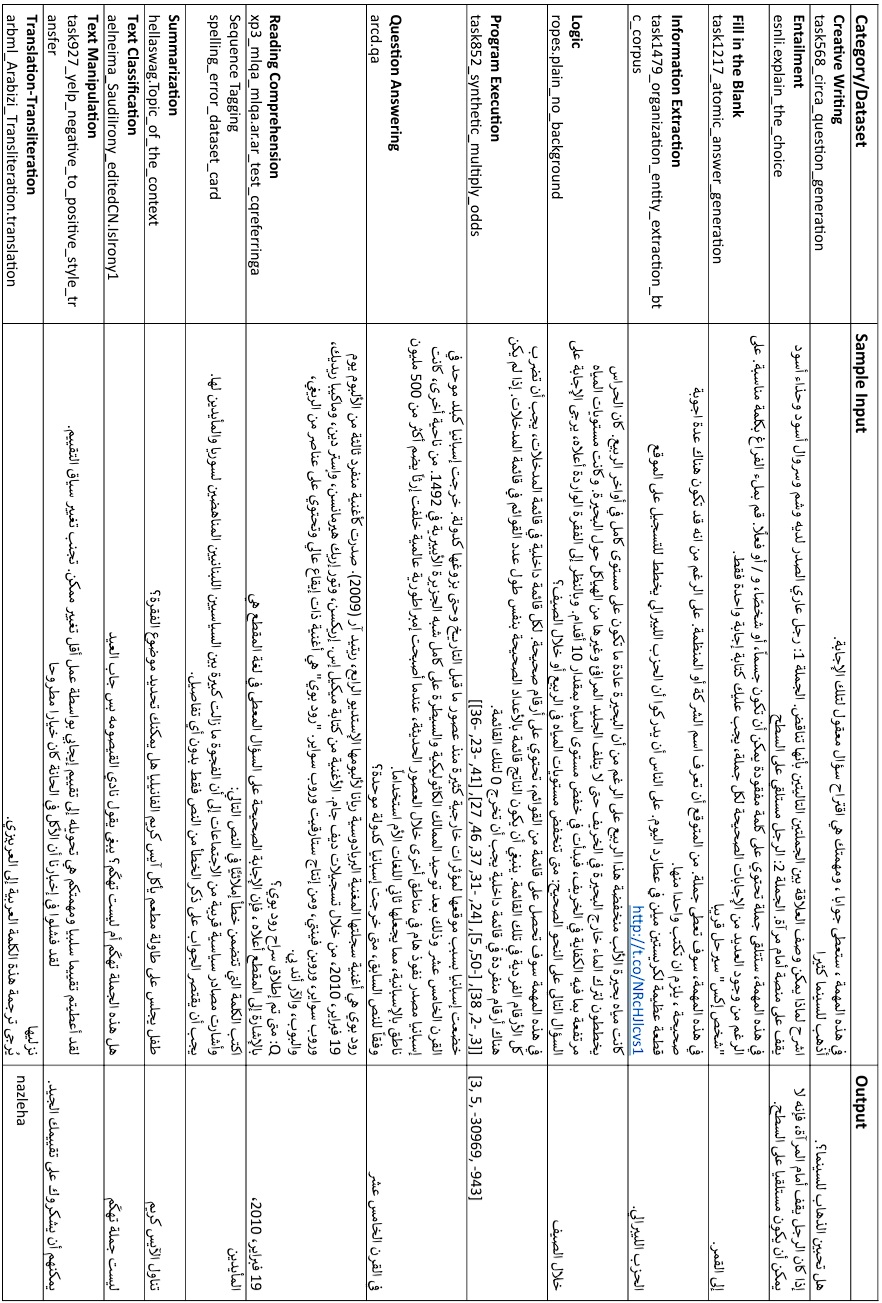}
    \caption{Samples from different categories. 
    }
    \label{table:samples}
\end{figure*}

Figure \ref{table:samples} shows examples of some prompts and responses for the different categories.

\section{Models}
\label{sec:app_models_details}

\subsection{Open-source Models}
\begin{itemize}[leftmargin=*]
    \item \textbf{AceGPT-v2-8B-Chat~\cite{huang-etal-2024-acegpt}:} A fine-tuned Arabic dialogue model based on LLaMA2, designed for chat-style interactions in Arabic.
    
    \item \textbf{Aragpt2-mega (1.5B)~\cite{huang-etal-2024-acegpt}:} A large-scale Arabic GPT-2 model designed for generating and understanding Arabic text.
    
    \item \textbf{c4ai-aya-expanse-32b~\cite{dang2024ayaexpansecombiningresearch}:} A multilingual large language model supporting 23 languages, including Arabic, with strong performance across diverse tasks.
    
    \item \textbf{Command R+ (104B):\footnote{\href{https://huggingface.co/CohereLabs/c4ai-command-r-plus}{https://huggingface.co/CohereLabs/c4ai-command-r-plus}}} A multilingual model optimized for retrieval-augmented generation (RAG), reasoning, and task completion, with general Arabic support.
    
    \item \textbf{Command-r7b 12-2024:\footnote{\href{https://huggingface.co/CohereLabs/c4ai-command-r7b-12-2024}{https://huggingface.co/CohereLabs/c4ai-command-r7b-12-2024}}} A compact and efficient version of the Command family of models, designed for general-purpose instruction following and language generation.
    
    \item \textbf{DeepSeek V3 (685B)\cite{deepseekai2024deepseekv3technicalreport}:} A multilingual Mixture-of-Experts model for reasoning, coding, and language understanding.
    
    \item \textbf{Gemma2 9B~\cite{team2024gemma}:} A multilingual language model from Google.
    
    \item \textbf{Jais-family 13b-chat~\cite{sengupta2023jais}:} A bilingual Arabic-English model trained on 395B tokens, optimized for long-sequence handling.
    
    \item \textbf{qwen-2.5 32b~\cite{qwen2}:} A high-capacity language model with strong performance in Chinese and English and expanding capabilities in other languages, including Arabic.
    
    \item \textbf{SILMA-9B Instruct-v1.0:\footnote{\href{https://huggingface.co/silma-ai/SILMA-9B-Instruct-v1.0}{https://huggingface.co/silma-ai/SILMA-9B-Instruct-v1.0}}} A 9-billion-parameter Arabic language model built on Google's Gemma architecture, fine-tuned for instruction-following tasks.
    
    \item \textbf{Yehia-7B preview:\footnote{\href{https://huggingface.co/Navid-AI/Yehia-7B-preview}{https://huggingface.co/Navid-AI/Yehia-7B-preview}}} A bilingual model designed for Arabic and English, capable of instruction-following and engaging in natural dialogue.

    \item \textbf{Fanar~\cite{fanar2024}:} It comes with two 7B and 9B parameter LLMs trained on nearly 1 trillion tokens. The models are designed to support both Arabic and English.
    
    \item \textbf{Mistral large :\footnote{\href{https://huggingface.co/mistralai/Mistral-Large-Instruct-2407}{https://huggingface.co/mistralai/Mistral-Large-Instruct-2407}}} A multilingual model with 123B parameters by Mistral AI.
    
    \item \textbf{DBRX-instruct (132B):\footnote{\href{https://huggingface.co/databricks/dbrx-instruct}{https://huggingface.co/databricks/dbrx-instruct}}} An instruction-tuned transformer developed by Databricks for high-quality reasoning and generation.

\end{itemize}

\subsection{Closed-Source Models}
\begin{itemize}[leftmargin=*]
    \item \textbf{Nuha v2 :\footnote{\href{https://nuha.ai/}{https://nuha.ai/}}}Nuha is an advanced, culture-aware AI assistant infused with pre-training and fine-tuning to understand Arabic nuances. With Nuha that is 40B parameter .
    
    \item \textbf{Iron Horse Gamma Velorum V5a:\footnote{\href{https://www.ironhorse.ai/}{https://www.ironhorse.ai/}}} A closed-source MoE model with 1.1T - 2.3T parameters based on the request. It supports more than 25 language. 
.
    \item \textbf{Amazon Nova Pro~\cite{intelligence2024amazon}:} A multilingual model by Amazon Bedrock designed for commercial applications.

    \item \textbf{Mistral-saba-latest (24B):\footnote{\href{https://mistral.ai/news/mistral-saba}{https://mistral.ai/news/mistral-saba}}} An Arabic fine-tuned variant of the Mistral model.

    \item \textbf{Grok-2-latest (314B MoE):} A closed-source model by xAI, designed for reasoning and factual recall.
    
    \item \textbf{Claude Sonnet 3.5:\footnote{\href{https://www.anthropic.com/news/claude-3-5-sonnet}{https://www.anthropic.com/news/claude-3-5-sonnet}}} A multilingual and instruction-capable model by Anthropic, estimated at over 130B parameters.

    \item \textbf{Gemini 2.0 Flash:} A lightweight variant of Gemini 2.0 optimized for speed and extended context.
    
    \item \textbf{GPT-4o~\cite{achiam2023gpt}:} OpenAI’s model supporting multimodal and multilingual input, including Arabic.

\end{itemize}

\section{Example LLM Outputs with the Same Meaning}
\label{appendix:example-outputs}
Consider the following Question Answering example where the correct answer is \<باريس> (Paris). While SILMA 9B just answered with \<باريس> only, the other model responses were much more verbose. Consider the answer of c4ai-aya-expanse-32b: \begin{scriptsize} \begin{RLtext} \<الجواب هو: باريس. وفقًا للفقرة، بدأت مرحلة تتابع الشعلة في فرنسا من باريس، حيث انعقدت في 7 أبريل، وبدأت على المستوى الأول من برج إيفل، وانتهت في ملعب تشارليتي.> \end{RLtext} \end{scriptsize} \begin{small} -- \textbf{Translation:} \textit{The answer is: Paris. According to the piece, the journey of the torch started in France from Paris on April 7 where it started from the first level of the Eiffel Tower and ended at the Charléty stadium}. \end{small}
BLEU is computed using the geometric mean of word unigram, bigram, trigram, and 4-gram precisions. Since many of the gold answers were short, resulting in no matching tri- and 4-grams, BLEU scores for many examples were zeros, despite the presence of matching unigrams and bigrams.

\section{Prompts for LLM-Based Evaluation}
Here is the prompt we used to extract the correct answer only from the LLM output:

\begin{lstlisting}[language={},style=pythonstyle,caption={Prompt for LLM-based answer extraction.},label={lst:prompt_for_answer}]
"""Given the following prompt: 

{prompt}

And the following automatically generated output:

{response}

Extract the answer from the automatically generated output ONLY WITHOUT any modification. Remove all non-related text from the answer. Do not put any additional text. If there are multiple answers, extract the first one only.
"""
\end{lstlisting}

Here is the prompt that we used for LLM as a judge:

\begin{lstlisting}[language={},style=pythonstyle,caption={LLM-as-a-Judge prompt.},label={box:llm-as-a-judge}]
You are an impartial and expert judge evaluating the quality of text generated by another AI model.
Your task is to score the generated output based on the original prompt and a provided ground truth answer, following a specific scoring rubric.
You will be provided with three pieces of information:
1.  The original prompt given to the generative model.
2.  The ground truth answer, representing the ideal or expected output.
3.  The actual output generated by the generative model.
Evaluate the generated output by comparing it to the ground truth, considering how well it addresses the original prompt.

Scoring Rubric:
*   Score 0: The automatically generated output is completely wrong, irrelevant, or unrelated to the prompt and ground truth.
*   Score 1: Poor answer. The output attempts to address the prompt but contains significant errors, is largely incomplete, or is difficult to understand. It shows little resemblance to the ground truth.
*   Score 2: Acceptable but different. The output is somewhat correct or addresses parts of the prompt reasonably well, but it differs significantly from the ground truth. It might be missing details present in the ground truth, include extra information not in the ground truth, or present the information in a substantially different structure or style, but it is still a valid (though not ideal) response to the prompt.
*   Score 3: Perfect or almost perfect. The output is accurate, complete, and closely matches the ground truth in content and style, effectively answering the original prompt. Minor differences in wording or formatting that do not affect the meaning or quality are acceptable for a score of 3.

Output Format:
Your output must be *only* a JSON object containing two keys:
1.  `score`: An integer between 0 and 3 based on the rubric above.
2.  `explanation`: A brief, concise string explaining *why* you assigned that score, referencing the differences or similarities between the generated output and the ground truth in the context of the prompt.

Example Output JSON:
{
  "score": 3,
  "explanation": "The generated output is accurate and complete, closely matching the ground truth."
}

[PROMPT]
{prompt}
[/PROMPT]

[GROUND TRUTH]
{reference answer}
[/GROUND TRUTH]

[GENERATED OUTPUT]
{response}
[/GENERATED OUTPUT]
\end{lstlisting}

\section{Human Evaluation Annotation Instructions}
\label{app:anno_instruct}

\begin{small}
\begin{RLtext}هل الإجابة صحيحة عند مقارنتها مع الإجابة الأصلية (0-3)؟\\
0: إجابة خاطئة تماما (لا تتطابق مع الإجابة الأصلية بأي شكل).\\
1: إجابة خاطئة جزئيا (تحتوي على بعض العناصر الصحيحة ولكن بها أخطاء جوهرية).\\
2: إجابة صحيحة جزئيا (تعكس بعض المعنى الصحيح ولكنها تفتقر إلى الدقة أو التفاصيل المهمة).\\
3: إجابة صحيحة تماما (متطابقة أو مكافئة للإجابة الأصلية دون أي أخطاء)
\end{RLtext}
\end{small}

\textbf{Translation of instructions:}\\
Is the answer correct when compared to the original answer (0-3)? \\
0: Completely wrong (does not match the original answer in any way).\\
1: Partially wrong (contains some correct elements but has significant errors).\\
2: Partially correct (conveys some correct meaning but lacks accuracy or important details).\\
3: Completely correct (identical or equivalent to the original answer with no errors).

\end{document}